\documentclass[conference]{IEEEtran}
\IEEEoverridecommandlockouts

\usepackage{cite}      % IEEE-style sorted and compressed citations
\usepackage{url}       % URL formatting for bibliography entries
\usepackage{balance}   % balance final-page columns

\usepackage{amsmath,amssymb,amsfonts}
\usepackage{mathtools}
\usepackage{bbm}   % blackboard-bold for digits/symbols, e.g., \mathbbm{1} for the indicator function
\usepackage{cases}
\usepackage{extarrows}
\usepackage{amsthm}

\usepackage{graphicx}

\usepackage[caption=false,font=footnotesize]{subfig}

\usepackage{booktabs}        % \toprule, \midrule, \bottomrule
\usepackage{threeparttable}  % table notes via tablenotes
\usepackage{array}           % custom column types
\usepackage{makecell}        % multi-line table headers
\usepackage{multirow}
\usepackage{tabularx}

\usepackage{xcolor}

\usepackage{algorithm}
\usepackage{algpseudocode}

\usepackage[english]{babel}
\usepackage{textcomp}
\addto\captionsenglish{\renewcommand{\figurename}{Fig.}}

\usepackage[colorlinks=true,
            linkcolor=blue,
            citecolor=blue,
            urlcolor=blue]{hyperref}

\usepackage{orcidlink}

\begin{document}

\title{TelecomGPT-R1: A Unified Open-Source Reasoner for the Telecom Stack}

\author{
	\IEEEauthorblockN{Bohao Wang\IEEEauthorrefmark{1}\IEEEauthorrefmark{2}, Chenwei Wu\IEEEauthorrefmark{4}, Haoyu Li\IEEEauthorrefmark{3}, Hang Zou\IEEEauthorrefmark{2}, Yu Tian\IEEEauthorrefmark{2}, Lina Bariah\IEEEauthorrefmark{2}, Li Wei\IEEEauthorrefmark{1}, Chongwen Huang\IEEEauthorrefmark{1},}
	\IEEEauthorblockN{Yongliang Shen\IEEEauthorrefmark{3}, Zhaoyang Zhang\IEEEauthorrefmark{1}, and M\'{e}rouane~Debbah\IEEEauthorrefmark{2}, \IEEEmembership{Fellow,~IEEE}}

	\IEEEauthorblockA{\IEEEauthorrefmark{1} College of Information Science and Electronic Engineering, Zhejiang University, 310027, Hangzhou, China}
	\IEEEauthorblockA{\IEEEauthorrefmark{2} Research Institute for Digital Future, Khalifa University, 127788, Abu Dhabi, UAE}
    \IEEEauthorblockA{\IEEEauthorrefmark{4} Department of Electrical Engineering and Computer Science, University of Michigan, Ann Arbor, MI 48109-2122, USA}
	\IEEEauthorblockA{\IEEEauthorrefmark{3} College of Computer Science and Technology, Zhejiang University, 310027, Hangzhou, China}
    Email: \{bohaowang, 3220105244, chongwenhuang, syl, ning\_ming\}@zju.edu.cn, chenweiw@umich.edu, \\
    \{bohao.wang, hang.zou, yu.tian, lina.bariah, merouane.debbah\}@ku.ac.ae, weili\_xd@163.com
}

\maketitle

\begin{abstract}
Telecommunications is a high-leverage domain for large language model (LLM)-based reasoning because routine engineering workflows require joint grounding in normative specifications, operational telemetry, vendor-specific fault evidence, and exact RF/network calculations. However, current LLM integration in telecom remains bottlenecked by a two-sided capability gap: generic reasoners often lack telecom-specific grounding, while domain-specific telecom LLMs remain limited in structured, multi-step reasoning. To bridge this gap, we release TelecomGPT-R1-9B, a unified open-source telecom reasoner that ranks top-performing on the GSMA open telco leaderboard. 
Specifically, we curate a 67{,}427-example supervised fine-tuning (SFT) corpus organized around four complementary reasoning axes: protocol, knowledge, modeling, and fault. The corpus is built from axis-matched public web sources and enhanced through axis-specific chain-of-thought (CoT) generation and prefix-continuation self-validation. Starting from Qwen3.5-9B, we further develop a two-stage post-training recipe. First, multi-teacher low-rank adaptation (LoRA)-based SFT injects telecom knowledge and induces axis-specific reasoning formats. Second, group relative policy optimization (GRPO), stabilized by decoupled clip and dynamic sampling policy optimization (DAPO), optimizes the policy using four axis-aligned binary verifier rewards. Across seven public telecom benchmarks, TelecomGPT-R1-9B ranks first among open-source telecom LLMs and achieves a seven-axis mean comparable to state-of-the-art closed-source frontier reasoners.
\end{abstract}

\begin{IEEEkeywords}
Large language models, telecommunications, post-training, supervised fine-tuning, reinforcement learning, verifiable rewards, GRPO, DAPO, universal reasoning.
\end{IEEEkeywords}

\section{Introduction}\label{sec:intro}

Large language models (LLMs) have advanced rapidly in reasoning and comprehension~\cite{jaech2024openai,guo2025deepseek}, offering great promise for the telecommunications domain. A capable telecom LLM could automate a wide range of engineering tasks, including standards interpretation, protocol analysis, log diagnosis, configuration inspection, wireless formula verification, and code-level troubleshooting. This broad application potential has motivated a growing body of telecom-oriented LLMs, benchmarks, datasets, and reasoning workflows~\cite{zou2025telecomgpt,maatouk2025teleqna}.

However, current telecom LLM integration remains far from day-to-day deployment. While existing LLMs perform well on static question answering (QA) tasks~\cite{maatouk2025teleqna}, real-world telecom engineering requires complex reasoning across diverse tasks and data modalities. For example, diagnosing a network failure may require an engineer to connect a 3rd generation partnership project (3GPP) procedure with an open radio access network (O-RAN) configuration table~\cite{gajjar2025oransight}, trace abnormal events in system logs, verify whether key performance indicator (KPI) changes match expected protocol behavior, and check whether the implementation or configuration violates a standard-defined constraint. Such workflows reveal a core challenge: telecom does not speak a single data language. A deployable telecom reasoner must therefore reason across heterogeneous source types, where standards, logs, tables, formulas, and code each require different evidence formats, reasoning patterns, and correctness criteria.

This gap is not yet solved by either general-purpose reasoners or existing telecom-domain LLMs. Frontier reasoning models such as GPT-5~\cite{singh2025openai} and Claude Opus 4.5~\cite{anthropic2025claudeopus45} demonstrate strong general reasoning, mathematical, and coding abilities. However, they lack domain knowledge and are not trained to reliably ground their reasoning in telecom-specific standards, counters, protocol procedures, or operational evidence. A model may produce a coherent-looking derivation while misreading structured radio access network (RAN) logs as free-form text~\cite{gajjar2025oransight}, confusing 3GPP working-group scopes, or hallucinating non-existent information elements, timers, counters, and protocol procedures. These failures reveal a mismatch between general reasoning ability and the dense, standards-constrained knowledge required for telecom engineering.

On the other hand, telecom-specialized models improve domain exposure, but they often focus on static QA, retrieval-assisted answering, or source-specific tasks rather than unified reasoning across standards, logs, tables, formulas, and code. TelecomGPT~\cite{zou2025telecomgpt} and Tele-LLMs~\cite{maatouk2026tele} adapt general-purpose LLMs to telecom through curated corpora, continued pre-training, instruction tuning, and task-oriented alignment, improving domain knowledge, telecom QA, classification, and related technical tasks. WirelessMathLM~\cite{li2025wirelessmathlm} and ORANSight-2.0~\cite{gajjar2025oransight} further demonstrate the value of reinforcement learning with verifiable rewards and retrieval-augmented instruction tuning for wireless mathematics and O-RAN understanding. However, these efforts remain centered on specific task families or evidence sources, and therefore leave open the problem of training one unified policy that can produce long, grounded reasoning chains across heterogeneous telecom evidence. In this work, we therefore study the important but underexplored question: \textit{\textbf{How can we build a unified telecom reasoning model that generalizes across heterogeneous telecom tasks and data sources?}}

To answer this, we release \textit{TelecomGPT-R1-9B}, the best telecom reasoning model to date. We first curate a 67{,}427-example supervised fine-tuning (SFT) corpus organized along four reasoning axes: protocol, knowledge, modeling, and fault. Each axis is populated from axis-matched public web sources and processed with axis-specific chain-of-thought (CoT) generation and prefix-continuation self-validation, so that each axis contributes reasoning traces aligned with its evidence structure. We adopt a two-stage post-training recipe using Qwen3.5-9B~\cite{qwen35} as the backbone. Multi-teacher low-rank adaptation (LoRA)-based SFT first installs telecom knowledge, response discipline, and axis-specific reasoning formats into the base model. We then perform reinforcement learning (RL) with decoupled clip and dynamic sampling policy optimization (DAPO)~\cite{yu2026dapo}, a variant of group relative policy optimization (GRPO), under per-axis binary verifier rewards. Specifically, we apply asymmetric trust-region clipping, dynamic sampling, token-level loss aggregation, and an SFT-anchored Kullback-Leibler (KL) regularizer to stabilize the RL training and facilitate convergence. Evaluated on the GSMA open telco leaderboard across seven public telecom benchmarks, TelecomGPT-R1-9B ranks \#1 among open-source telecom LLMs and remains comparable with state-of-the-art (SOTA) closed-source frontier reasoners. We further conduct an empirical study of post-training design choices for improving unified telecom reasoning capabilities and identify practical lessons for data curation, SFT, and RL algorithm design, supporting future works in this domain.

In summary, our contributions are three-fold:
\begin{itemize}
\item \textbf{New Formulation.} We formulate telecom reasoning as a unified policy across heterogeneous tasks and data modalities, where a model must reason across standards, logs, tables, formulas, code, and configuration evidence under source-specific correctness criteria.
\item \textbf{New Data and Model.} We publicly release the TelecomGPT-R1-9B model weights. TelecomGPT-R1-9B ranks \#1 among open-source telecom LLMs on the GSMA open telco leaderboard across seven public benchmarks.
\item \textbf{New Findings.} We identify practical findings for post-training telecom reasoners, including the importance of domain SFT before RL, source-matched CoT generation, dynamic sampling, verifier design, and teacher diversity.
\end{itemize}

% Fig. 1 (radar) moved to Section III Experimental Evaluation — see 4_simulation_conf.tex
\section{Method}
\label{sec:method}

\begin{figure*}[!t]
    \centering
    \includegraphics[width=0.85\textwidth]{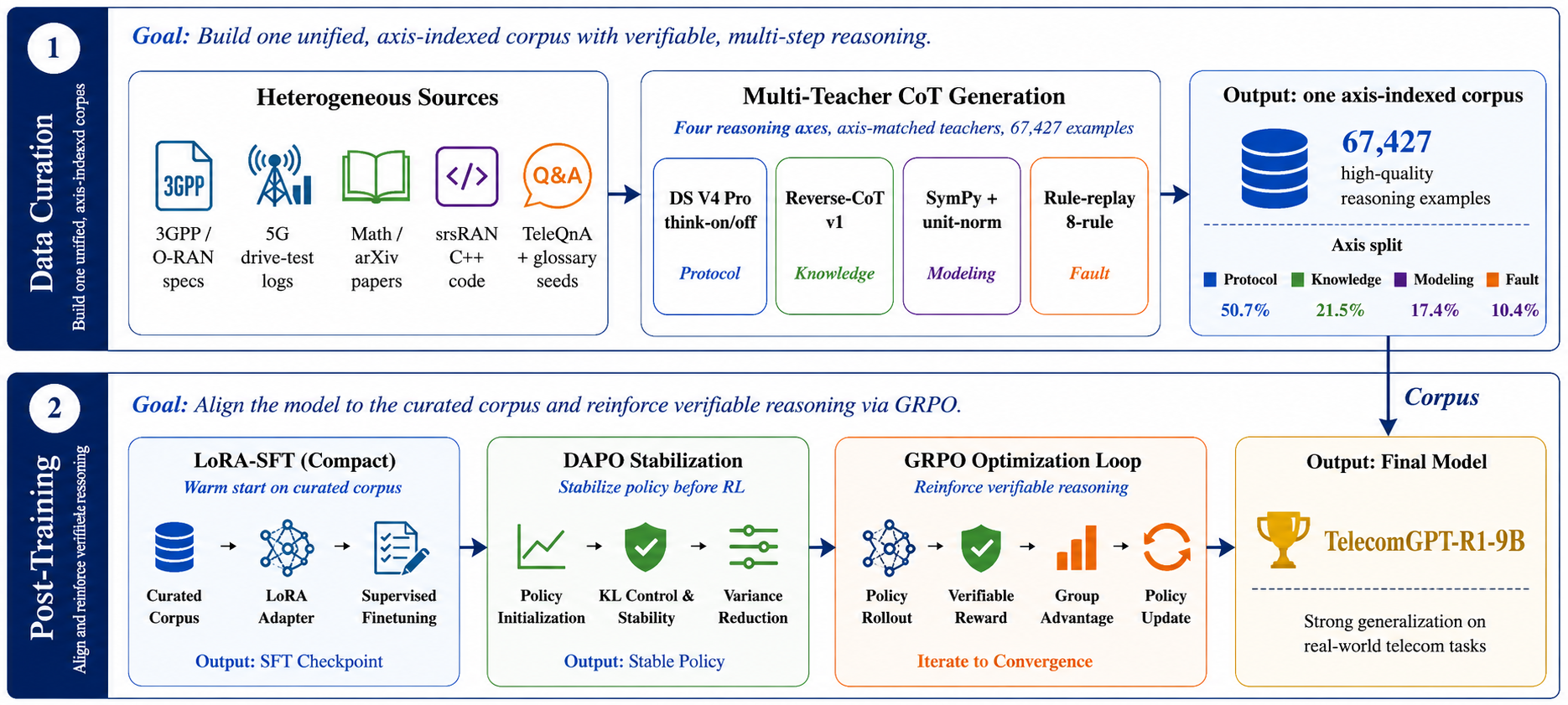}
    \caption{Two-frame TelecomGPT-R1-9B recipe. Frame~\textcircled{1} curates the 67{,}427-example corpus via axis-matched multi-teacher CoT generation across the four reasoning axes (protocol/knowledge/modeling/fault). Frame~\textcircled{2} post-trains Qwen3.5-9B with LoRA-SFT followed by a DAPO-stabilized GRPO loop on per-axis binary verifier rewards (asymmetric clip, dynamic sampling, token-level loss, KL anchor).}
    \label{fig:recipe}
\end{figure*}

\subsection{Data Curation}
\label{sec:method:data}

Telecom reasoning spans heterogeneous evidence formats, reasoning styles, and verification criteria. We therefore construct the training corpus through four steps: defining reasoning axes, collecting source materials, synthesizing or extracting questions, and generating verified CoT traces.

\paragraph{Reasoning axes}
We first organize telecom reasoning into four axes: $\mathrm{protocol}$, $\mathrm{knowledge}$, $\mathrm{modeling}$, and $\mathrm{fault}$. The protocol axis targets reasoning over normative 3GPP and O-RAN specifications. The knowledge axis targets factual and terminological question answering over operator, vendor, and general telecom knowledge. The modeling axis targets RF, queueing, networking, mathematical, and code-grounded computation. The fault axis targets root-cause analysis from operational RAN evidence.

\paragraph{Source collection}
For each axis, we collect publicly accessible sources that expose the axis's native evidence structure. Protocol examples are derived from public 3GPP specification PDFs, 3GPP technical reports, and O-RAN alliance specifications, accounting for 50.7\% of the corpus. Knowledge examples are derived from public standard documents, research papers, and telecom glossary entries from operator documentation, accounting for 21.5\%. Modeling examples are derived from the srsRAN project C++ repository, telecom-engineering textbooks and lecture notes, and arXiv telecom mathematics papers, accounting for 17.4\%. Fault examples follow the synthetic data-generation pipeline of~\cite{wang2025towards}, which couples user-side records with engineering-parameter tables to infer the root cause of downlink throughput below 600 Mbps in a 5G RAN, accounting for the remaining 10.4\%. Together, these sources yield a 67{,}427-example SFT corpus, summarized in Fig.~\ref{fig:recipe}.

\paragraph{Question synthesis and extraction}
We convert each source family into telecom reasoning question--answer pairs. For specification-based protocol data, we parse public 3GPP and O-RAN PDFs, project document and working-group labels, and synthesize questions that require clause-level or procedure-level understanding. For implementation-grounded modeling data, we ground questions in abstract syntax tree and doxygen symbols from srsRAN~\cite{gomez2016srslte}, then synthesize hierarchy-aware multiple-choice questions (MCQ) with distractors drawn from nearby modules, functions, or protocol concepts. For mathematical modeling data, we parse textbooks, lecture notes, and arXiv papers with MinerU~\cite{wang2024mineru}, mask formulas or derivation targets, and generate questions that require recovering or applying the missing computation. For knowledge data, we convert general telecom facts and glossary entries into factual and terminological QA examples. For table reasoning, we construct a 3GPP-table sub-corpus using 371 real 3GPP tables and synthesize table-grounded distractors. For fault analysis, we build on the diagnostic setting introduced by~\cite{wang2025towards}, but expand its original coarse rule set into a rule replay system that separates calculation, threshold checking, and root-cause assignment.

\paragraph{CoT generation and verification}
Finally, we generate source-matched reasoning traces rather than relying on a single generic CoT prompt. Protocol and knowledge questions use teacher-LLM rationales filtered by self-validation. Mathematical examples use python-grounded CoT traces that are re-executed and checked by symbolic equivalence with unit tolerance. Fault-analysis examples use deterministic rule replay. The teacher LLM computes relevant throughput and radio indicators, records which of the diagnostic rules fire, and derives the final root-cause label in the structured \verb|[Calculation]/[Rules]/[Answer]| format. This ensures that fault-analysis rationales are exactly consistent with the diagnostic rule engine rather than merely teacher-written explanations. Across all axes, generated traces are further filtered by prefix-continuation self-validation, where the rationale is replayed as a forced prefix and the gold answer must remain re-derivable from the continuation.

After generation, all examples pass through shared normalization: multi-pass verification, augmentation, leakage filtering, difficulty stratification, and style mixing, before being normalized to a standard \verb|{system, user, assistant}| schema with per-row axis and source tags. Each axis is paired with a binary verifier $V^{(a)}$: protocol and knowledge use \verb|\boxed{L}|/\verb|ANSWER: $L$| option matching; modeling uses symbolic-equivalence checking with unit tolerance plus boxed-letter MCQ answering; fault uses deterministic rule-replay. The same verifier family supplies the RL rewards in Section~\ref{sec:method:posttrain}, yielding a single ablatable interface for axis re-weighting, CoT-generator replacement, and leakage-policy analysis.

\subsection{Post-Training: SFT and DAPO}
\label{sec:method:posttrain}

We post-train Qwen3.5-9B in two stages. Stage 1 performs knowledge installation and reasoning-format induction through LoRA-SFT. Let $\mathcal{D}=\{(x^{(n)}, y^{(n)})\}_{n=1}^{N}$ denote the curated trace set, and let $\pi_\theta$ denote the LoRA-adapted policy with adapter rank $r$ and scaling $\alpha=2r$. We minimize the token-level negative log-likelihood:
\begin{equation}
\label{eq:sft}
\mathcal{L}_{\mathrm{SFT}}(\theta) = -\,\mathbb{E}_{(x,y)\sim\mathcal{D}}\!\left[\frac{1}{|y|}\sum_{t=1}^{|y|} \log \pi_\theta\!\left(y_t \mid x, y_{<t}\right)\right].
\end{equation}
This stage exposes the base model to all four axes under a single interleaved training stream, thereby injecting telecom knowledge, answer discipline, and axis-specific CoT formats into one policy.

Stage 2 optimizes reasoning trajectories with a DAPO-style GRPO with KL regularization~\cite{guo2025deepseek,yu2026dapo}. For each prompt $x$, we sample a group of $G$ rollouts $\{o_i\}_{i=1}^{G}$ from the current policy. Let $a(x)$ denote the axis label of $x$, and let
$\rho_{i,t}(\theta)=\pi_\theta(o_{i,t}\!\mid\!x,o_{i,<t})/
\pi_{\theta_{\mathrm{old}}}(o_{i,t}\!\mid\!x,o_{i,<t})$
be the per-token importance ratio. We maximize the following objective
\begin{equation}
\label{eq:grpo}
\begin{aligned}
\mathcal{J}(\theta) ={} &
\mathbb{E}_{(x,\{o_i\})\in\mathcal{G}_{\mathrm{train}}}\!
\bigg[
\tfrac{1}{\sum_{i}|o_i|}
\sum_{i,t}
\min\!\Big(
\rho_{i,t}\hat{A}_{i,t}, \\
& \quad
\mathrm{clip}(\rho_{i,t}, 1{-}\varepsilon_\ell, 1{+}\varepsilon_h)\,\hat{A}_{i,t}
\Big)
\bigg] \\
& - \beta\, D_{\mathrm{KL}}(\pi_\theta\,\|\,\pi_{\mathrm{ref}}).
\end{aligned}
\end{equation}

where $\pi_{\mathrm{ref}}$ is the SFT reference policy. With $\eta>0$ for numerical stability, the group-relative advantage is
\begin{equation}
\label{eq:adv}
    \hat{A}_i = 
\frac{R(o_i) - \mathrm{mean}_{j=1}^{G}\, R(o_j)}
{\mathrm{std}_{j=1}^{G}\, R(o_j) + \eta}
\end{equation}

with the per-axis binary verifier reward
\begin{equation}
\label{eq:reward}
\begin{aligned}
R^{(a)}(o) ={} & \mathbbm{1}\!\left[V^{(a)}(o, y^\star) = 1\right], \\
& a \in \{\mathrm{protocol},\, \mathrm{knowledge},\, \mathrm{modeling},\, \mathrm{fault}\}.
\end{aligned}
\end{equation}
For a rollout group associated with prompt $x$, we set $R(o_i)=R^{(a(x))}(o_i)$.

Because all rewards are binary, many rollout groups carry little policy-gradient signal. We therefore apply dynamic sampling and retain only groups with non-degenerate reward distributions:
\begin{equation}
\label{eq:dynsampling}
\mathcal{G}_{\mathrm{train}} = \Big\{ (x, \{o_i\}_{i=1}^{G}) : 0 < \tfrac{1}{G}\!\sum_{i=1}^{G} R(o_i) < 1 \Big\}.
\end{equation}
This filter removes uniformly correct and uniformly incorrect groups before the policy update, focusing optimization on prompts where the current policy exhibits recoverable uncertainty. We further use asymmetric clipping with $\varepsilon_\ell=0.20$ and $\varepsilon_h=0.28$, token-level loss aggregation as in Eq.~\eqref{eq:grpo}, and an SFT-anchored KL regularizer with $\beta=0.001$. The asymmetric clip preserves useful exploratory updates in long 3GPP and fault-analysis traces. Different from the original DAPO paper, we find the KL anchor is beneficial for maintaining structured formats installed during SFT, including \verb|[Calculation]/[Rules]/[Answer]| and tail-JSON outputs. Overall, SFT installs telecom knowledge and axis-specific reasoning formats, while verifier-driven RL improves trajectory selection under a single unified policy. The complete recipe is summarized in Fig.~\ref{fig:recipe}.

% --- Conference Experimental Evaluation: F1-F4 + Tab 1 + figures live in 4_simulation_conf.tex ---
\section{Experimental Evaluation}
\label{sec:sim}

We evaluate TelecomGPT-R1-9B against frontier baselines on the GSMA open telco leaderboard, spanning seven benchmarks: 3GPP-TSG~\cite{gsma2026open_telco_evals}, ORANBench~\cite{gajjar2025oran}, srsRANBench~\cite{gajjar2025oransight}, TeleLogs~\cite{sana2025reasoning}, TeleMath~\cite{colle2026telemath}, TeleQnA~\cite{maatouk2025teleqna}, and TeleTables~\cite{ezzakri2025teletables}. We start from the Qwen3.5-9B base model~\cite{qwen35} and run all training on an $8{\times}$H200 node. Supervised fine-tuning applies QLoRA on all attention and MLP projections of the LLM with rank $r=128$, learning rate $10^{-4}$, and two epochs over the 67{,}427-example corpus. Verifier-driven RL then runs 100 DAPO update steps from the SFT initialization with asymmetric clip $(\varepsilon_\ell, \varepsilon_h)=(0.20, 0.28)$ and KL coefficient $\beta=0.001$.

\begin{figure}[t]
    \centering
    \includegraphics[width=0.98\columnwidth]{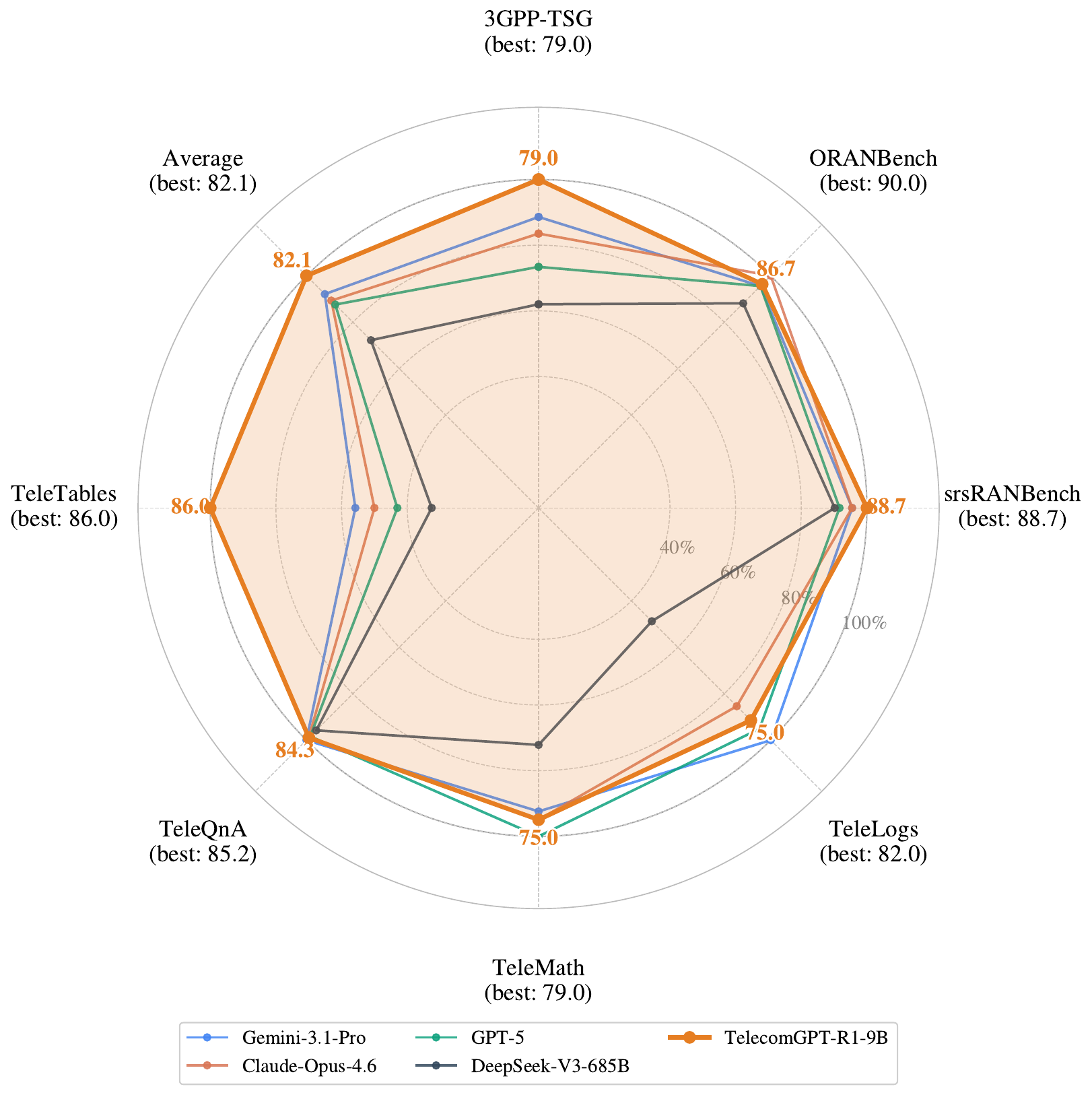}
    \caption{TelecomGPT-R1-9B vs.\ frontier baselines on the GSMA open telco leaderboard.}
    \label{fig:radar}
\end{figure}

With 9B parameters, our model attains a mean score of $82.1\%$, a new SOTA among open-source telecom LLMs and within the closed-source frontier tier on the seven-axis aggregate. Fig.~\ref{fig:radar} visualizes this result. Each spoke is one of the seven benchmarks plus the average, normalized by the per-axis leaderboard best. TelecomGPT-R1-9B traces the outer edge among open-source models, exceeding the current best open-source generalist DeepSeek-V3, a $685$B-parameter model that reaches a $59.3\%$ seven-axis mean, by $+22.8$\,pp at roughly $1/75$ of the parameter count, and stays within the closed-source frontier tier alongside GPT-5 at $71.9\%$, Claude-Opus-4.6 at $73.3\%$, and Gemini-3.1-Pro at $75.6\%$. Table~\ref{tab:gsma} reports per-axis accuracy for the three post-training stages SFT, SFT+GRPO, and SFT+DAPO from a shared 9B Qwen3.5 LoRA-SFT init; we release the SFT+DAPO checkpoint as TelecomGPT-R1-9B. We attribute the result to the full recipe rather than to DAPO alone. The 67,427-example axis-indexed corpus, source-matched CoT generation, SFT with teacher diversity, and the DAPO-stabilized GRPO objective each contribute. We distill these contributions into the following four findings.

\begin{table}[t]
\renewcommand{\arraystretch}{1.15}
\caption{Per-axis accuracy on the GSMA open telco leaderboard; best per column in \textbf{bold}.}
\label{tab:gsma}
\centering
\setlength{\tabcolsep}{1.5pt}
\scriptsize
\begin{tabular}{l|c c c c c c c c}
\hline\hline
Stage & Avg & 3GPP. & ORAN. & srsRAN & TeleLogs & TeleMath & TeleQnA & TeleTab. \\
\hline
SFT       & 75.2          & 71.0          & \textbf{88.0} & 84.7          & 42.0          & 68.0          & 83.5          & \textbf{89.0} \\
SFT+GRPO  & 77.2          & 70.0          & 84.7          & 87.3          & 51.0          & \textbf{76.0} & \textbf{84.3}          & 87.0 \\
SFT+DAPO  & \textbf{82.1} & \textbf{79.0} & 86.7          & \textbf{88.7} & \textbf{75.0} & 75.0          & \textbf{84.3} & 86.0 \\
\hline
\end{tabular}
\end{table}

\subsection{Domain knowledge is the bottleneck}
\label{sec:sim:f1}

\begin{figure}[t]
    \centering
    \includegraphics[width=0.98\columnwidth]{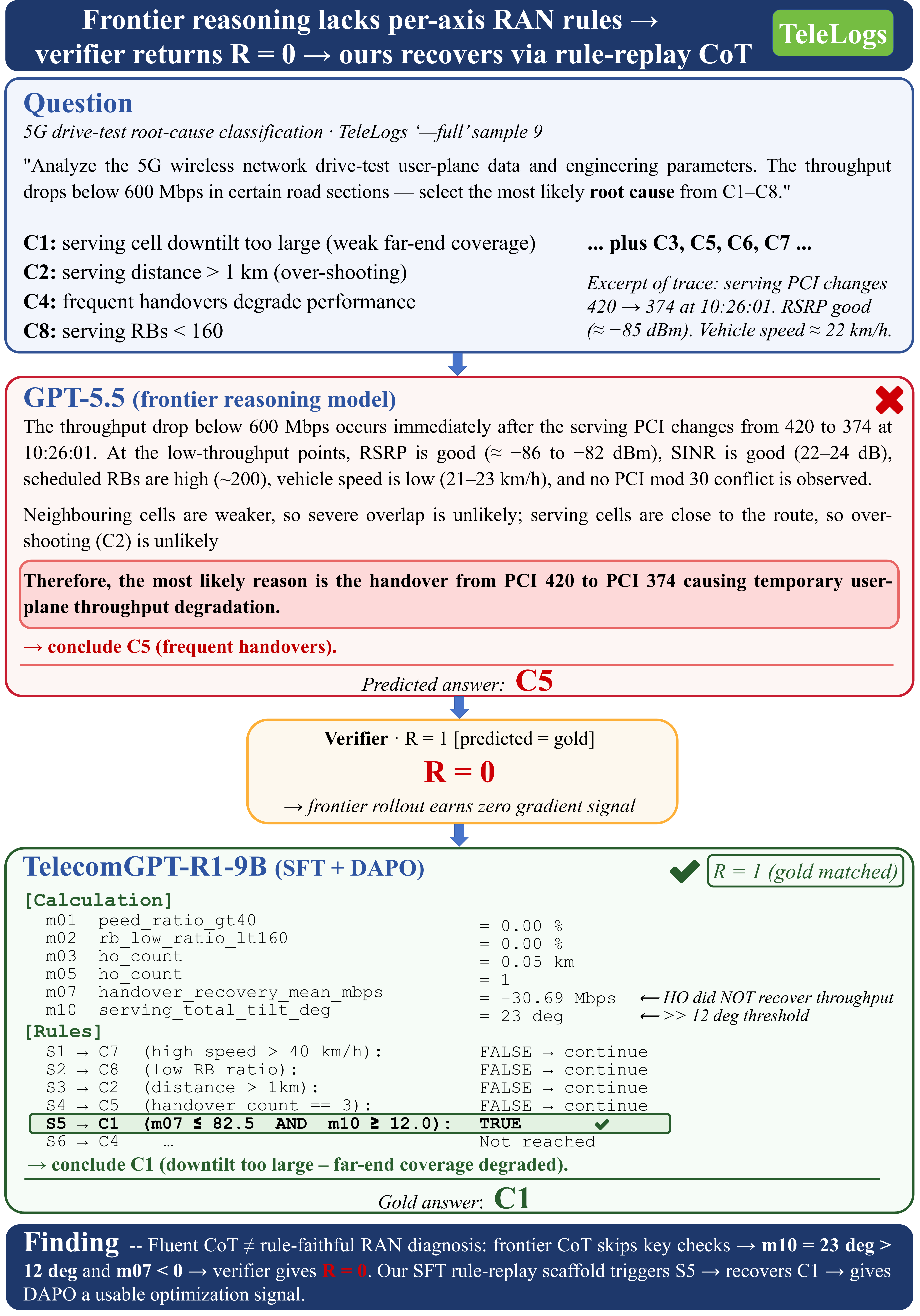}
    \caption{Rule-blind frontier vs.\ verifier-aligned CoT on TeleLogs.}
    \label{fig:wrong_fact}
\end{figure}

The first lesson is that SFT is not merely format alignment. It installs the domain facts and source conventions RL needs before verifier rewards become useful. When the base model lacks the underlying telecom fact, the rollout is fluent but wrong, the verifier returns $R=0$ across the group, and DAPO has no learning signal. Fig.~\ref{fig:wrong_fact} illustrates this on TeleLogs, where GPT-5.5 narrates a rule-blind CoT, observes a PCI handover at the throughput-drop boundary, and picks C5 without checking the downtilt threshold or the handover-recovery sign that define C1. Our model walks the SFT-installed rule-replay scaffold, computes $m_{07}=-30.69$\,Mbps and $m_{10}=23^{\circ}$, hits rule S5, and recovers C1. This failure mode dominates errors on TeleLogs and 3GPP-TSG. Parameter count is not a substitute for the per-axis verifier rules SFT installs.

The implication is a clean attribution. SFT carries the knowledge load, RL then improves trajectory selection, output discipline, and
robustness on axes where the verifier provides non-degenerate signals. Empirically, as quantified in Table~\ref{tab:gsma}, the SFT-only checkpoint reaches a mean of $75.2\%$, with DAPO adding $+6.9$\,pp on top to reach $82.1\%$. The largest DAPO gains land on the under-saturated axes: TeleLogs jumps from $42.0\%$ to $75.0\%$ ($+33$\,pp), and 3GPP-TSG rises from $71.0\%$ to $79.0\%$ ($+8$\,pp). The other three axes ORANBench, TeleQnA, and TeleTables are SFT-saturated and fluctuate within $1\text{--}3$\,pp around their SFT ceilings.

\subsection{Source-matched CoT generators matter}
\label{sec:sim:f2}

\begin{table*}[t]
\centering
\caption{CoT-regime ablation on TeleMath and TeleLogs.}
\label{tab:cot_regime}
\footnotesize
\setlength{\tabcolsep}{6pt}
\renewcommand{\arraystretch}{1.15}
\begin{tabular}{l|l|l|c c}
\hline\hline
Stage & CoT corpus & Design keywords & TeleMath & TeleLogs \\
\hline
\multirow{3}{*}{\emph{SFT only}}
 & NoCoT                            & Answer-only target, no reasoning trace                    & 42.2                       & 35.4 \\
 & BadCoT                           & Single generic teacher, source-blind CoT                  & 72.0                       & 59.0 \\
 & Multi-source, ours               & Axis-matched 4-quadrant generators, multi-teacher mixture & 68.0                       & 42.0 \\
\hline
\multirow{2}{*}{\emph{DAPO + 100 RL steps}}
 & BadCoT SFT init                  & Single generic teacher, source-blind CoT                  & 65.2                       & 72.9 \\
 & Multi-source SFT init, ours & Axis-matched 4-quadrant generators, multi-teacher mixture & 75.0                       & 75.0 \\
\hline
\end{tabular}
\end{table*}

% We partition the corpus (67{,}427 rows organized along the four reasoning axes) into a 4-quadrant CoT-generation taxonomy, matching each generator to the epistemic structure of its source: (Q1) teacher-LLM distillation via DS V4 Pro thinking-on/off for TeleQnA and the 3GPP narrative axis; (Q2) symbol-grounded derivation using \texttt{sympy} with unit normalization for TeleMath; (Q3) Python-grounded execution with executable traces and log parsing for TeleLogs and network modeling; and (Q4) rule-replay verification combining deterministic spec lookup with reverse CoT for TeleTables and fault classification. Corpus mass is allocated 50.7\,/\,21.5\,/\,17.4\,/\,10.4\,\% across the four quadrants.

The second lesson is that a single generic CoT generator is unreliable across telecom sources, because what counts as evidence differs across axes. We therefore use the source-matched CoT generators, pairing each axis with a verifier-aligned generator rather than a single generic teacher prompt, so that arithmetic mistakes, hallucinated table evidence, and skipped diagnostic thresholds are caught before SFT.

Table~\ref{tab:cot_regime} isolates the effect of source-matched CoT generators on the two axes most sensitive to CoT design. TeleMath benefits from symbol-grounded derivation and TeleLogs benefits from deterministic rule-replay traces. The NoCoT baseline trains on answer-only targets and drops sharply on both axes because the model cannot allocate reasoning capacity to the symbolic and rule-driven evidence each task requires. The BadCoT row is more revealing. Its SFT-only score reaches 72.0\,/\,59.0 on TeleMath\,/\,TeleLogs and initially appears competitive with our multi-source SFT at 68.0\,/\,42.0. Yet BadCoT compounds poorly under DAPO and lands at 65.2\,/\,72.9, well below our DAPO checkpoint at 75.0\,/\,75.0. The mechanism is that SFT installs the reasoning scaffold that DAPO subsequently polishes through verifier feedback. Our multi-source SFT encodes a verifier-aligned reasoning structure with room for the binary verifier to identify and reinforce correct intermediate steps, leaving substantial headroom for DAPO to compound. The BadCoT prior, by contrast, encodes a shallow answer-format pattern that lacks any structural surface for verifier rewards to grip, so DAPO updates degenerate and the policy quickly stops improving.

\subsection{DAPO outperforms GRPO under heterogeneous source training}
\label{sec:sim:f3}

The third lesson is that unified telecom RL is not a single homogeneous optimization problem, because different sources saturate at different speeds so static sampling wastes rollouts on easy prompts and undertrains difficult ones. Table~\ref{tab:gsma} shows the post-SFT checkpoint is already near-saturated on the MCQ-shaped axes ORANBench, TeleQnA, and TeleTables but weak on TeleLogs and underdeveloped on 3GPP-TSG, so these two are the axes RL must lift. Fig.~\ref{fig:converge_a} traces per-axis accuracy over 100 update steps and confirms DAPO outperforms GRPO most strongly on TeleLogs and 3GPP-TSG. We omit srsRANBench and TeleTables because they overlap with ORANBench and TeleQnA at the saturated band. On the MCQ-shaped axes the two policies converge within $\sim$1--2\,pp of the SFT ceiling because the binary letter-match verifier returns near-degenerate gradient signal regardless of sampling policy, so DAPO's gain over GRPO comes from dynamic sampling re-allocating rollouts to non-saturated TeleLogs and 3GPP-TSG prompts rather than from a uniform per-axis lift.

% ===== FIG 5 REMOVED 2026-06-10: mean Δ-acc trajectory merged into the Mean panel of Fig.~\ref{fig:converge_a}. =====

\subsection{Teacher diversity boosts SFT quality}
\label{sec:sim:f4}

\begin{figure}[t]
    \centering
    \includegraphics[width=0.98\columnwidth]{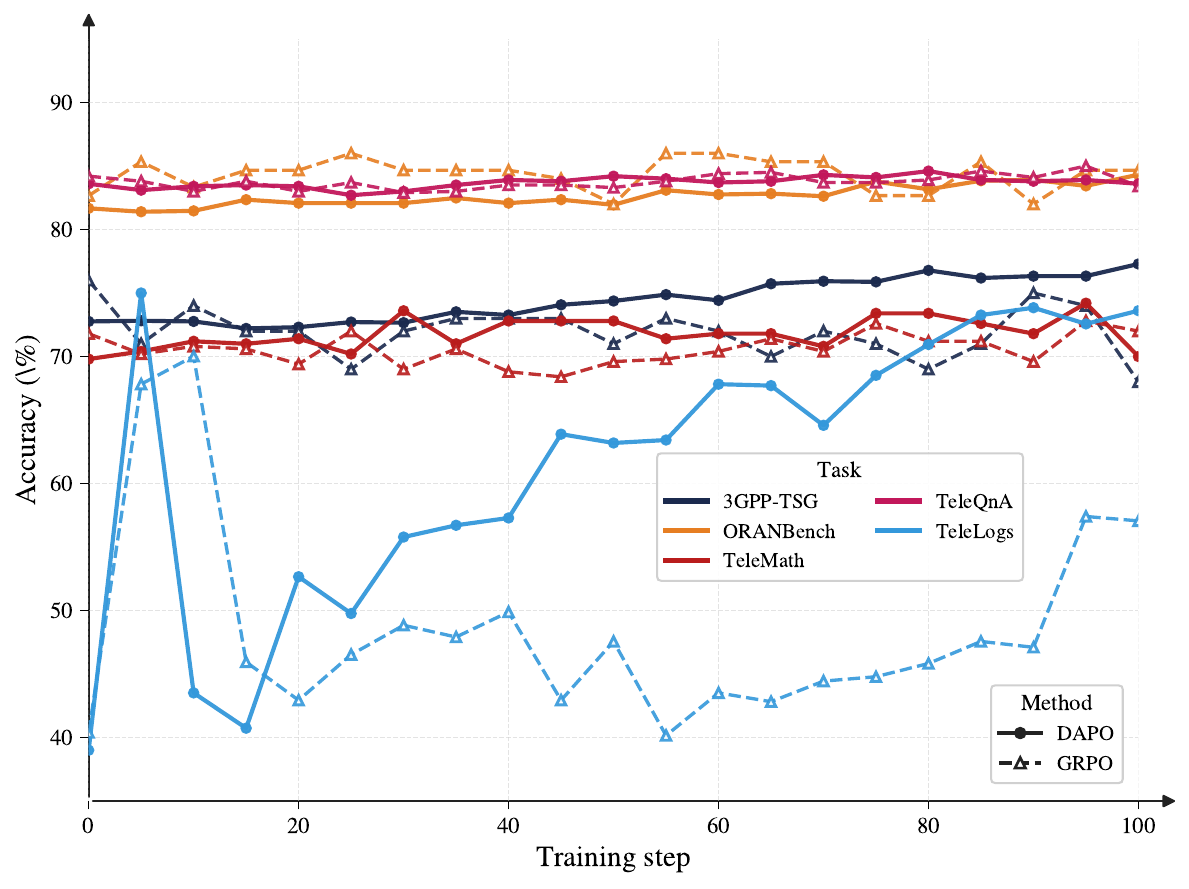}
    \caption{Per-task DAPO vs.\ GRPO trajectories.}
    \label{fig:converge_a}
\end{figure}

The fourth lesson is that teacher diversity improves the SFT corpus, since a single teacher tends to overfit to a narrow explanation style. We address this through a multi-model teacher mixture spanning four of the six SFT generators, complemented by a small share of general long-reasoning chain-of-thought for cross-domain reflection. Table~\ref{tab:cot_regime} grounds this attribution. Holding corpus size and target axes fixed, the gap between single-teacher BadCoT and our multi-teacher multi-source pipeline on TeleMath and TeleLogs attributes cleanly to teacher diversity rather than to additional rollouts or verifier feedback. On TeleMath and TeleLogs, multi-source SFT init with DAPO reaches $75.0/75.0$ against $65.2/72.9$ from BadCoT-init DAPO, showing that teacher diversity primes the policy for RL exploitation single-teacher BadCoT cannot achieve. Fig.~\ref{fig:converge_a} further traces per-axis training dynamics, where TeleLogs and 3GPP-TSG attain the largest DAPO gains under dense rule-based and long-tail standard signals while the MCQ axes ORANBench, TeleQnA, and TeleTables plateau within a narrow band around their SFT ceilings.

\section{Conclusion}
\label{sec:conclusion}

We present TelecomGPT-R1-9B, a unified open-source telecom reasoner that grounds general reasoning capability in telecom-specific evidence. Specifically, we curate a 67{,}427-example axis-indexed corpus over protocol, knowledge, modeling, and fault reasoning, then post-train Qwen3.5-9B with multi-teacher LoRA-SFT followed by DAPO-stabilized GRPO under axis-aligned binary verifier rewards. On the GSMA open telco leaderboard, TelecomGPT-R1-9B ranks first among open-source telecom LLMs and remains within the closed-source frontier tier on the seven-axis mean, showing that structured data curation, verifier-coupled training, and unified policy optimization convert heterogeneous telecom evidence into transferable reasoning capability. Future work will scale the backbone to larger models and broaden the corpus across additional telecom evidence sources. We will also continue refining the post-training framework to further improve reasoning capability on heterogeneous telecom tasks.

% \clearpage
% IEEEtran's \thebibliography already wraps in \footnotesize — do NOT add \small
% here, it would override IEEE's standard reference size.
\makeatletter
\let\@oldthebibliography\thebibliography
\renewcommand{\thebibliography}[1]{%
  \@oldthebibliography{#1}%
  \setlength{\itemsep}{0pt plus 0.3pt}%
  \setlength{\parsep}{0pt}%
  \setlength{\topsep}{0pt}%
}
\makeatother
\balance
\bibliographystyle{IEEEtran}
\bibliography{bib}

\end{document}